\documentclass[10pt,twocolumn,letterpaper]{article}

\usepackage{cvpr}
\usepackage{times}
\usepackage{epsfig}
\usepackage{graphicx}
\usepackage{amsmath}
\usepackage{amssymb}
\usepackage{balance}
\usepackage{multirow}
\usepackage{booktabs}
\usepackage[small,tight]{subfigure}

\usepackage[pagebackref=true,breaklinks=true,letterpaper=true,colorlinks,bookmarks=false]{hyperref}

\cvprfinalcopy 

\def\cvprPaperID{****} 

\ifcvprfinal\fi
\begin{document}

\title{UC Merced Submission to the ActivityNet Challenge 2016}

\author{Yi Zhu \quad Shawn Newsam\\
University of California, Merced, USA \\
{\tt\small \{yzhu25,snewsam\}@ucmerced.edu}
\and
Zaikun Xu\\
University of Lugano, Switzerland\\
{\tt\small xuz@usi.ch}
}

\maketitle

\begin{abstract}
This notebook paper describes our system for the untrimmed classification task in the ActivityNet challenge $2016$. We investigate multiple state-of-the-art approaches for action recognition in long, untrimmed videos. We exploit hand-crafted motion boundary histogram features as well feature activations from deep networks such as VGG16, GoogLeNet, and C3D. These features are separately fed to linear, one-versus-rest support vector machine classifiers to produce confidence scores for each action class. These predictions are then fused along with the softmax scores of the recent ultra-deep ResNet-101 using weighted averaging. 
\end{abstract}

\section{Introduction}
\label{sec:introduction}
Human action recognition in video is a fundamental problem in computer vision due to its increasing importance for a range of applications such as video recommendation and search, video highlighting, video surveillance, human-robot interaction, human skill evaluation, etc.

The ActivityNet challenge \cite{activityNet} is a large scale benchmark designed to stimulate research on human activity understanding in user generated videos. This challenge consists of two tasks on 200 activity categories: (a) untrimmed classification and (b) detection. We focus on the former which involves predicting the activities present in a long video. Accounting for YouTube blocks and deleted videos, we downloaded $9942$ training, $4874$ validation, and $5001$ test videos. 

\begin{figure}[t]
	\centering
	\includegraphics[width=1.0\linewidth,trim=0 0 0 0,clip]{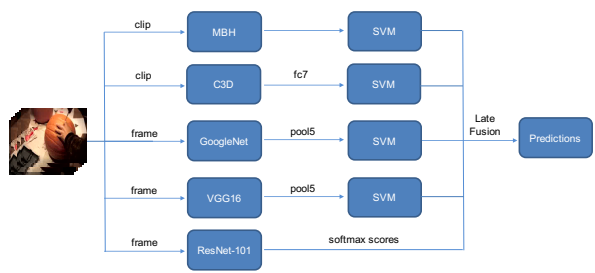}
	\vspace{-16ex}
	\caption{\textbf{Multi-stream framework}. We combine five modules using late fusion to obtain the final prediction scores. The MBH module is hand-crafted, while the rest are based on deep networks. For ResNet-101, we directly use the softmax scores since this performs better than using the extracted features.}
	\label{fig:workflow}
	\vspace{-2ex}
\end{figure}

\section{Recognition Framework}
\label{sec:methodology}
In this section, we present our multi-stream action recognition framework based on: (i) Fisher vector encoded MBH features, (ii) C3D \textsf{fc7} features, (iii) GoogLeNet \textsf{pool5} features, (iv) VGG16 \textsf{pool5} features, and (v) ResNet-101 softmax scores. The first two modules are clip-based while the last three are frame-based. An overview of the framework can be found in Fig. \ref{fig:workflow}. 

\subsection{MBH Features}
Improved dense trajectories (IDT) \cite{idtfWang2013} are state-of-the-art hand-crafted features for modeling temporal information in videos, and the motion boundary histogram (MBH) features are the best performing component of the IDT features. We use the provided\footnote{The MBH features are provided by the organizers.} Fisher vector encoded MBH features \cite{FV2013,compact_fisher_action_oneata_iccv13}, whose dimension is $65536$ for each video, to train a linear, one-versus-rest support vector machine (SVM) classifier. We fix the SVM hyper-parameter $C$ to $100$ \cite{liblinear}.

\subsection{C3D}
In \cite{c3d2015}, the authors show that 2D ConvNets ``forget'' the temporal information in the input signal after each convolution. They therefore propose 3D ConvNets, which analyze sets of contiguous video frames organized as clips, and show its effectiveness at learning spatio-temporal features in video volume data analysis problems.

We therefore adopt \textsf{fc7} features\footnote{The C3D extracted \textsf{fc7} features are provided by the organizers.} extracted from a pre-trained C3D model as an additional signal. The network is not fine-tuned on the ActivityNet challenge dataset. The inputs to the C3D model are $16$ frame clips with $50\%$ overlap and the outputs are $4096$ dimension feature activations. These features are reduced to $500$ dimensions using PCA. Average pooling is used to combine the clip-level features for a single video. Finally, a linear, one-versus-rest SVM is trained with $C$ set to $1$.

\subsection{GoogLeNet}
We also use features\footnote{The GoogLeNet extracted \textsf{pool5} features are provided by the organizers.} extracted from the \textsf{pool5} layer of a Google inception net (GoogLeNet) \cite{MettesICMR16}. This network is an enhanced version of \cite{Szegedy_2015_CVPR} which utilizes a reorganized hierarchy of the complete ImageNet dataset \cite{imagenet_cvpr09}. The features are frame-based and have dimension $1024$. They are mean-pooled across all frames in a video followed by L1-normalization to obtain a video-level representation. Again, a linear, one-versus-rest SVM is trained with $C$ set to $1$.

\subsection{VGG16}
VGG16 \cite{vgg16192015} is a popular deep architecture that demonstrated good performance on action recognition in \cite{wanggoodpractice2015} using a two-stream \cite{twostream2014} pipeline. We only employ the spatial stream. We use the pre-trained VGG16 model for initialization and fine-tune it on the challenge dataset. During fine-tuning, we perform $60$K iterations with learning rate $10^{-4}$, $30$K iterations with $10^{-5}$, and $30$K iterations with $10^{-6}$. Momentum and weight decay are set to $0.9$ and $5 \times 10^{-4}$.

We adopt the latent concept descriptor (LCD) encoding method in \cite{LCDXu2015} to encode the \textsf{pool5} layer of our fine-tuned VGG16 model, followed by VLAD encoding \cite{VLAD}. We reduce the dimensions of the \textsf{pool5} features from $512$ to $256$ using PCA. The number of centers in VLAD encoding is set to $256$ and we use VLAD-k with $k$ set to $5$. The encoded features are then power- and intra-normalized. The resulting per-frame features have dimension $65536$ which are mean-pooled to obtain a video-level representation. A linear, one-versus-rest SVM is trained with $C$ set to $1$.

\subsection{ResNet-101}
Residual learning \cite{residual_cvpr16} has recently become an effective method to construct ultra-deep networks for object recognition and detection. We extend it here to action recognition. We adopt the pre-trained $101$-layer model for initialization and fine-tune it on the ActivityNet video data. The learning rate is $10^{-4}$ for the first two epochs, $10^{-5}$ for the following two epochs, and $10^{-6}$ for the last epoch.  Momentum and weight decay are set to $0.9$ and $10^{-4}$.
 
We also investigated using features extracted from last convolutional module, whose dimension is $2048$, to train an SVM, similar to our other modules. This, however, performs $3.3\%$ worse on the validation set than using the softmax scores. 

\section{Experiment Results}
\label{sec:experiments}
Given a test video, we uniformly sample $25$ frames to extract the frame-level feature activations and perform mean-pooling to obtain the final video representation.

Late fusion iteratively combines pairs of prediction scores. First, the outputs of two modules are combined in a weighted fashion where the scores of the more accurate module are weighted twice that of the less accurate one. Additional scores are then combined with this in a similar fashion. After late fusion, we adopt a Multi-class Iterative Re-ranking (MIR) method \cite{rankingReranking2015} to re-rank the predictions of classifiers based on the difficulty scale of the videos.
Table \ref{tab:validation} shows our experimental results on the validation set of the ActivityNet challenge 2016. 

\begin{table}
	\begin{center}
		\resizebox{0.9\columnwidth}{!}{%
			\begin{tabular}{ c | c  }
				\toprule
				Model								& 	Top-1 Accuracy	\\
				\hline		
				(i) MBH												 & 	$57.32\%$ 	\\		
				(ii) C3D \textsf{fc7}							 &   $60.04\%$ 		\\	
				(iii) GoogLeNet \textsf{pool5} 				&   $67.13\%$ 	\\
				(iv) VGG16$^{\ast}$	\textsf{pool5} 	   &   $63.19\%$ 	\\
				(v) ResNet-101$^{\ast}$ 					   &   $71.81\%$ 	\\
				\hline
				(i) + (ii)								 &   $62.78\%$ 	\\		
				(i) + (iii)								 &   $69.40\%$ 		\\	
				(i) + (iv)								&   $68.79\%$ 	\\
				(ii) + (iii) 								 &   $68.11\%$ 	\\
				(ii) + (iv) 								&   $64.35\%$ 	\\
				(iii) + (iv) 								&   $68.56\%$ 	\\
				(ii) + (iii) + (iv) 						&   $69.09\%$ 	\\
				(i) + (v)									&   $73.05\%$ 	\\
				(ii) + (iii) + (iv) + (v) 				  &   $73.56\%$ 	\\
				(i) + (iii) + (iv) + (v) 			 			&   $74.68\%$ 	\\
				(i) + (ii) + (iii) + (iv) + (v) 			 &   $75.14\%$ 	\\
				\bottomrule
			\end{tabular}
		}
		\vspace{4ex}
		\caption{Action recognition results on the validation set of the ActivityNet challenge 2016. All performances are reported using top-1 accuracy. Top: Single module performances. Bottom: Fused module performances. $^{\ast}$ indicates the network is fine-tuned on the challenge dataset. \label{tab:validation}}
		\vspace{-3ex}
	\end{center}
\end{table} 

We can see from the top part of Table \ref{tab:validation} that the residual network achieves the best performance among all modules. It is $14.5\%$ better than the state-of-the-art hand-crafted MBH features and outperforms the other deep networks. 

The bottom part of Table \ref{tab:validation} shows the performances of various module combinations. We observe that combinations that only include deep networks are generally not as effective as combinations that include the MBH features. Although the MBH features perform the worst alone, they are orthogonal to the deep learning based approaches. This may be attributed to MBH being effective at capturing low-level motion features while the deep networks model high-level information related to static appearance. The MBH features and the deep networks are thus quite complementary. When fusing all modules, our system achieves a validation accuracy of $75.14\%$.
%

We also investigate incorporating action proposals generated by \cite{fast_temporal_action_proposal_fabian_cvpr16} during prediction. Instead of uniformly sampling $25$ frames across the video, we sample $25$ frames from the action proposals. The intuition is that these action proposals have a higher probability of containing action frames, so that the average pooling of these frames should lead to higher recognition accuracy. However, this turns out to perform worse than uniform sampling.

\section{Submission Details}
We use both the training and validation data as the training set for our submissions. Note, though, that the implementation details and parameter settings remain the same as when we use only the training data to train. We do not use the test data for training or parameter tuning. 

\begin{table}
	\begin{center}
		\resizebox{0.8\columnwidth}{!}{%
			\begin{tabular}{ c | c | c | c }
				\toprule
				Submission								& 	mAP    & Top-1 Accuracy	& Top-3 Accuracy	 \\
				\hline		
				Run 1								& 	$68.00\%$ 	& 	$66.16\%$ & 	$83.36\%$ \\		
				Run 2								&   $75.98\%$ 	& 	$72.48\%$ & 	$87.54\%$ 	\\	
				Run 3								&   $79.41\%$   & 	$76.17\%$ & 	$90.19\%$ 	\\
				Run 4 								&   $81.64\%$   & 	$77.74\%$ & 	$90.93\%$ 	\\
				Run 5 								&   $83.1\%$     & 	$78.44\%$  & 	$91.07\%$ 	\\
				\bottomrule
			\end{tabular}
		}
		\vspace{2ex}
		\caption{Action recognition results on the test set of the ActivityNet challenge 2016. \label{tab:test}}
		\vspace{-4ex}
	\end{center}
\end{table} 

We submit five runs to the evaluation server, and the performance for each run is shown in Table \ref{tab:test}. Our runs are as follows:
\begin{itemize}
	\item Run 1: VGG16
	\item Run 2: VGG16 + MBH
	\item Run 3: VGG16 + MBH + ResNet-101
	\item Run 4: VGG16 + MBH + ResNet-101 + GoogLeNet 
	\item Run 5: VGG16 + MBH + ResNet-101 + GoogLeNet + C3D 
\end{itemize}

\section{Conclusion}
We show that the ultra-deep architecture of ResNet is indeed helpful in learning discriminative features for complex tasks, such as human activity understanding. In addition, although hand-crafted MBH features achieve the lowest accuracy alone, they are complementary to approaches based on deep networks. Finally, the combination of all modules using late fusion gives the best performance.

\section*{Acknowledgement}
We gratefully acknowledge the support of NVIDIA Corporation through the donation of the Titan X GPU used in this work. This work was funded in part by a National Science Foundation CAREER grant, \#IIS-1150115.

{\small
\bibliographystyle{ieee}
\bibliography{egbib_cvpr}
}

\end{document}